\pdfoutput=1

\documentclass[11pt,a4paper]{article}
\usepackage{url}
\usepackage[hyperref]{emnlp2020}

\usepackage{times}
\usepackage{latexsym}

\usepackage{url}
\usepackage{booktabs}       
\usepackage{amsfonts}       
\usepackage{nicefrac}       
\usepackage{microtype}      
\usepackage{algorithmic}
\usepackage{algorithm}
\usepackage{wrapfig}
\usepackage{lipsum}
\usepackage{xspace}
\usepackage{graphicx} 
\usepackage{subfigure} 
\usepackage[skip=0pt]{caption}
\usepackage{paralist}
\usepackage{float}
\usepackage{tabularx}
\usepackage{multirow}
\usepackage{diagbox}
\usepackage{tikz}
\usepackage{booktabs}
\usepackage{array, makecell} 
\usepackage{natbib}
\usepackage{amsmath,amssymb}
\usepackage[inline]{enumitem}
\usepackage{acronym}

\usepackage[belowskip=-10pt,aboveskip=4pt]{caption}

\DeclareMathOperator*{\argmax}{arg\,max}

\newcounter{todocnt}

\acrodef{RL}{Reinforcement Learning}
\acrodef{PL}{Policy Learning}
\acrodef{DRL}{Deep Reinforcement Learning}
\acrodef{IRL}{inverse reinforcement learning}
\acrodef{SERP}{search engine result page}
\acrodef{IR}{Information Retrieval}
\acrodef{MDP}{Markov Decision Process}
\acrodef{MaxEnt-IRL}{Maximum Entropy Inverse Reinforcement Learning}
\acrodef{DM-IRL}{Distance Minimization Inverse Reinforcement Learning}
\acrodef{MMI}{Maximum Mutual Information} 
\acrodef{DNN}{Deep Neural Networks}
\acrodef{RNN}{Recurrent Neural Networks}
\acrodef{MLP}{Multilayer Perceptron}
\acrodef{GRU}{Gated Recurrent Net}
\acrodef{TDSs}{Task-oriented Dialogue Systems}
\acrodef{TDS}{Task-oriented Dialogue System}
\acrodef{LU}{language understanding}
\acrodef{DM}{Dialogue Management}
\acrodef{NLG}{natural language generation}
\acrodef{DST}{Dialogue State Tracker}
\acrodef{SOTA}{state-of-the-art}
\acrodef{NLU}{Natural Language Understanding}

\aclfinalcopy
\renewcommand{\aclpaperid}{2573}

\title{Dialogue Policy Learning: Supervised, Adversarial and Reinforcement Learning}
\title{Rethinking Supervised Learning and Reinforcement Learning \\ in Dialogue Policy Learning}
\title{Rethinking Supervised Learning and Reinforcement Learning \\ in Task-Oriented Dialogue Systems}

\author{
Ziming Li \textsuperscript{1},
Julia Kiseleva \textsuperscript{2},
Maarten de Rijke \textsuperscript{1}\\
\textsuperscript{1}University of Amsterdam,
\textsuperscript{2}Microsoft Research\\
\{z.li,m.derijke@@uva.nl\},
julia.kiseleva@microsoft.com
}

\date{}

\begin{document}
\maketitle

\begin{abstract}
Dialogue policy learning for \ac{TDSs} has enjoyed great progress recently mostly through employing \ac{RL} methods. However, these approaches have become very sophisticated. It is time to re-evaluate it. Are we really making progress developing dialogue agents only based on \ac{RL}? We demonstrate how (1)~traditional supervised learning together with (2)~a simulator-free adversarial learning method can be used to achieve performance comparable to \ac{SOTA} \ac{RL}-based methods. 
First, we introduce a simple dialogue action decoder to predict the appropriate actions. Then, the traditional multi-label classification solution for dialogue policy learning is extended by adding dense layers to improve the dialogue agent performance. Finally, we employ the Gumbel-Softmax estimator to alternatively train the dialogue agent and the dialogue reward model without using \ac{RL}. 
Based on our extensive experimentation, we can conclude the proposed methods can achieve more stable and higher performance with fewer efforts, such as the domain knowledge required to design a user simulator and the intractable parameter tuning in reinforcement learning. Our main goal is not to beat \ac{RL} with supervised learning, but to demonstrate the value of rethinking the role of \ac{RL} and supervised learning in optimizing \ac{TDSs}.
\end{abstract}

\setlength{\abovedisplayskip}{3pt}
\setlength{\belowdisplayskip}{3pt}

\section{Introduction}
\label{sec:introduction}

The aim of dialogue policies in \ac{TDS} is to select appropriate actions at each time step according to the current context of the conversation and user feedback~\citep{chen2017survey}.
In early work, dialogue policies were manually designed as a set of rules that map the dialogue context to a corresponding system action~\citep{weizenbaum1966eliza}.
The ability of rule-based solutions is limited by the domain complexity and task scalability. Moreover, the design and maintenance of these rules require a lot of effort and domain knowledge. 

Due to recent advantages in deep learning and the availability of labeled conversational datasets, \emph{supervised learning} can be employed for dialogue policy training to overcome the disadvantages of rule-based systems. 
The downside of the supervised learning approach is that the dialogues observed in the datasets are unlikely to represent all possible conversation scenarios; in some extreme cases, the required conversational dataset cannot be collected or acquiring it might cost-prohibitive. 

The success of \ac{RL} in other areas holds promises for dialogue \ac{PL}~\citep{williams2007partially}.
Using \ac{RL} techniques, we can train dialogue policies and optimize automatically, from scratch and utilizing interactions with users~\citep{gavsic2014gaussian,su2017sample}. 
In \ac{RL}-based solutions, the dialogue system takes actions that are controlled by the dialogue policy, and user feedback (the \emph{reward signal}), which is provided when the dialogue is finished, is utilized to adjust the initial policy~\citep{peng2018deep,williams2017hybrid,dhingra2016towards}. 
In practice, reward signals are not always available and may be inconsistent~\citep{su2016line}. 
As it is not practical to ask for explicit user feedback for each dialogue during policy training, different strategies have been proposed to design a rule-based user simulator along with a reward function that can approximate the real \emph{reward function} which exists only in each user's mind. 
Designing an appropriate user simulator and accurate reward function requires strong domain knowledge. 
This process has the same disadvantages as rule-based dialog systems~\citep{walker1997paradise}. 
The difference is that rule-based approaches to system design meet this problem at the dialogue agent side while rule-based user simulators need to solve it at the environment side.

If the task is simple and easy to solve, why not just build a rule-based system rather than a user-simulator that is then used with \ac{RL} techniques to train the dialogue system, where more uncontrollable factors are involved? 
And if the task domain is complex and hard to solve, is it easier to design and maintain a complicated rule-based user simulator than to build a rule-based dialogue agent?
Supervised learning methods do not suffer from these issues but require labeled conversational data; in some exceptional cases, if the data cannot be collected for privacy reasons, \ac{RL} is the solution. However, collecting labeled data is feasible for many applications~\citep{williams2014dialog,weston2015towards,budzianowski2018multiwoz}.
Therefore in this work seek to answer the following research question:
\emph{Are we really making progress in \ac{TDSs} focusing purely on advancing \ac{RL}-based methods?}

To address this question, we introduce three dialogue \ac{PL} methods which do not require a user simulator.
The proposed methods can achieve comparable or even higher performance compared to \ac{SOTA} \ac{RL} methods. 
The first method utilizes an action decoder to predict dialogue combinations. 
The second method regards the dialogue \ac{PL} task as a multi-label classification problem. 
Unlike previous work, we assign a dense layer to each action label in the action space.
Based on the second method, we propose an adversarial learning method for dialogue \ac{PL} without utilizing \ac{RL}. 
To backpropagate the loss from the reward model to the policy model, we utilize the Gumbel-Softmax to connect the policy model and the reward model in our third method. 
We compare our methods with \ac{RL} and adversarial \ac{RL} based dialogue training solutions to show how we can achieve comparable performance without a utilizing costly user simulator.

To summarize, our contributions are:
\begin{itemize}[nosep,leftmargin=*]
\item A dialogue action decoder to learn the dialogue policy with supervised learning.
\item A multi-label classification solution to learn the dialogue policy.
\item A simulation-free adversarial learning method to improve the performance of dialogue agents.
\item Achieving \ac{SOTA} performance in dialogue \ac{PL} with fewer efforts and costs compare to existing \ac{RL}-based solutions.
\end{itemize}

\section{Related Work}
\label{sec:rel_work}
A number of \ac{RL} methods, including Q-learning~\citep{peng2017composite, lipton2018bbq,li2017end,Su2018D3Q,li2020guided} and policy gradient methods \citep{dhingra2016towards,williams2017hybrid}, have been applied to optimize dialogue policies by interacting with real users or user simulators. \ac{RL} methods help the dialogue agent is able to explore contexts that may not exist in previously observed data. 
A key component in \ac{RL} is the quality of the reward signal used to update the agent policy. 
Most existing \ac{RL}-based methods require access to a reward signal based on user feedback or a pre-defined one if feedback loop is not possible. 
Besides, designing a good reward function and a realistic user simulator is not easy as it typically requires strong domain knowledge, which is similar to the problem that rule-base methods meet. 
\citet{peng2018adversarial} propose to utilize adversarial loss as an extra critic in addition to the main reward function based on task completion. 
Inspired by the success of adversarial training in other NLP tasks, \citet{liu2018adversarial} propose to learn dialogue rewards directly from dialogue samples, where a dialogue agent and a dialogue discriminator are trained jointly. 
Following the success of inverse reinforcement learning (IRL) in different domains,
\citet{takanobu2019guided} employ \emph{adversarial IRL} to train the dialogue agent. 
They replace the discriminator in GAIL~\citep{gan_imitation} with a reward function with a specific architecture. 
The learned reward function can provide a stable reward signal and adversarial training can benefit from high quality feedback.

Compared to existing \ac{RL} based methods, we propose strategy that can eliminate designing a user simulator and sensitive parameter-tuning process while bringing a significant performance improvement with respect to a number of metrics. The absence of user simulators involved will largely reduce the required domain knowledge and supervised learning can lead to robust agent performance.

\section{Multi-Domain Dialogue Agent}
\label{sec:method}

\noindent
\textbf{\ac{DST}} 
In a standard \ac{TDS} pipeline, the rule-based \ac{DST} is deployed to keep track of information emerging in interactions between users and the dialogue agent. 
The output from the \ac{NLU} module is fed to the \ac{DST} to extract information, including informable slots about the constraints from users and requestable slots that indicate what users inquire about. In our setup, the dialogue agents and user-simulators are interacting through predefined dialogue actions therefore no \ac{NLU} is involved. Besides, a belief vector is maintained and updated for each slot in every domain.

\noindent
\textbf{Dialogue state}
We formulate a structured state representation $s_t$ according to the information resulting from the \ac{DST} at time step $t$.
There are $4$ main types of information in the final representation: (1)~corresponding to the embedded results of returned entities for a query, 
(2)~the last user action, 
(3)~the last system action, and 
(4)~the belief state from the rule-based state tracker. 
The final state representation $s$ is a vector of $553$ bits.

\noindent
\textbf{Dialogue action}
We regard the dialogue response problem as a multi-label prediction task, where in the same dialogue turn, several atomic dialogue actions can be covered and combined at the same moment. 
In the action space, each action is a concatenation of domain name, action type and slot name, e.g.\ \textit{`attraction-inform-address'},
which we call an \emph{atomic action}\footnote{there are $166$ atomic actions in total in the action space}. 
\citet{lee2019convlab} proposes that the action space covers both the atomic action space and the top-k most frequent atomic action combinations in the dataset and then the dialogue \ac{PL} task can be regarded as a single label classification task. 
However, the expressive power of the dialogue agent is limited and it is beneficial if the agent can learn the action structure from the data and this could lead to more flexible and powerful system responses. 

\begin{figure}[t]
\centering
   \includegraphics[clip, width=0.65\columnwidth]{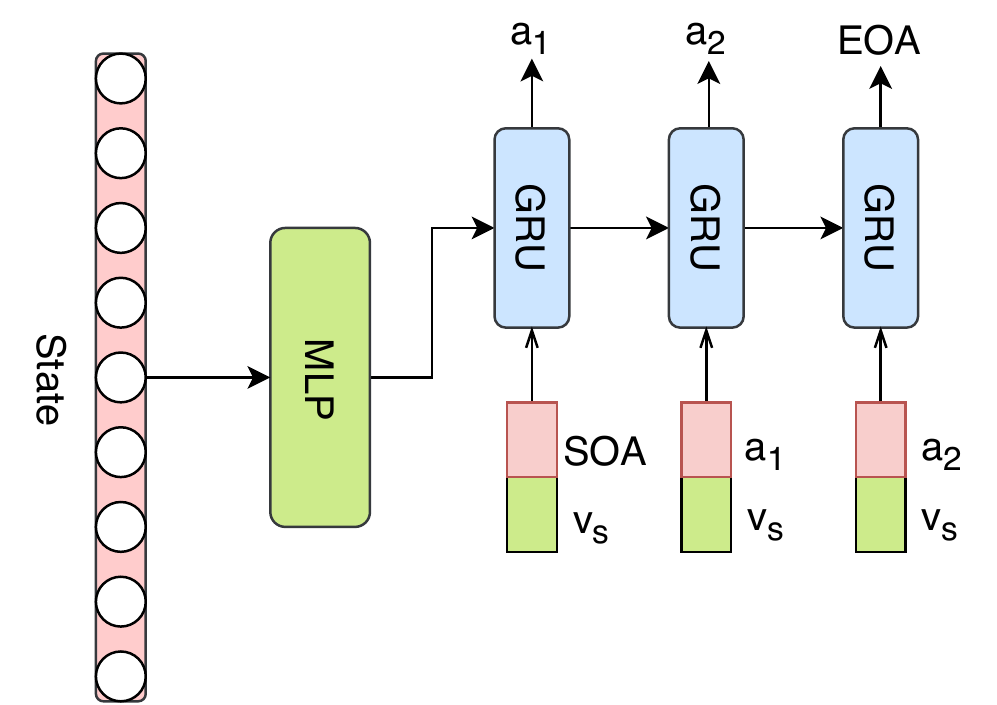}
   \caption{Architecture to approximate a dialogue policy with an action decoder\footnote{\textit{SOA} and \textit{EOA} are special actions corresponding to the starting signal and ending signal respectively}.}
   \label{fig:mlp_rnn}
     \vspace*{-1\baselineskip}
\end{figure}

\section{Dialogue Policy Learning (PL)} 
\acused{PL}
\subsection{\ac{PL} as a sequential decision process}
Different atomic dialogue actions contained in the same response are usually related to each other.
To fully make use of information contained in co-occurrence dependencies, we decompose the multi-label classification task in dialogue \ac{PL} as follows. Assuming the system response consists of two atomic actions, \textit{`hotel-inform-address'} and \textit{`hotel-inform-phone'}, the model takes the dialogue state as input and predict the atomic actions sequentially. The path could be described as either \textit{`hotel-inform-address'} $\rightarrow$ \textit{`hotel-inform-phone'} or \textit{`hotel-inform-phone'} $\rightarrow$ \textit{`hotel-inform-address'}. Before the training stage, the relative order of all the atomic actions will be predefined and fixed. 
Following this solution, we apply a GRU-based~\citep{cho2014learning} decoder to model the conditional dependency between the actions in one single turn as shown in Figure~\ref{fig:mlp_rnn}. 

The proposed model first extracts state features $v_s$ by feeding the raw state input $s$ to an \ac{MLP}.
In the next state, the state representation $v_s$ will be used as the initial hidden state $h_0$ of action decoder $GRU$. 
To avoid information loss during decoding, the input to the action decoder is: 
\begin{equation}
input_t = embedding(a_{t-1}) \oplus v_s.
\end{equation}
The starting input $input_0$ is the concatenation of starting action \textit{SOA} and state representation $v_s$. $a_{t-1}$ denotes the dialogue action in the prediction path at time step $t-1$ and \emph{embedding(a)} returns the action embedding of the given action \emph{a}. In the next steps, actions will be generated consecutively according to:
\begin{equation}
o_t, h_t = GRU (input_t, h_{t-1}),
\end{equation}
where $o_t$ is the output of the action decoder. 
We use cross-entropy to train the action decoder together with the \ac{MLP} for feature extracting.
We use beam-search to find the most appropriate action path. 

\subsection{\ac{PL} with adversarial learning}	
\label{section:pladv}
Next, we introduce an adversarial learning solution, \emph{DiaAdv}, to train the dialogue policy without a user simulator along with a dialogue discriminator.
Feedback from the discriminator is used as a reward signal to push the policy model to interact with users in a way that is indistinguishable from how a human agent completes the task. However, since the output of the dialogue policy is a set of discrete dialogue actions, it is difficult to pass the gradient update from the discriminator to the policy model. To cross this barrier, we propose to utilize the Gumbel-Softmax function~\citep{jang2016categorical} to link the discriminator to the generator. Next, we will give a brief introduction about the dialogue policy model and the dialogue discriminator. Afterwards, we will show how we can utilize Gumbel-Softmax to backpropagate the gradient.

\begin{figure}[t]
\centering
   \includegraphics[clip, width=\columnwidth]{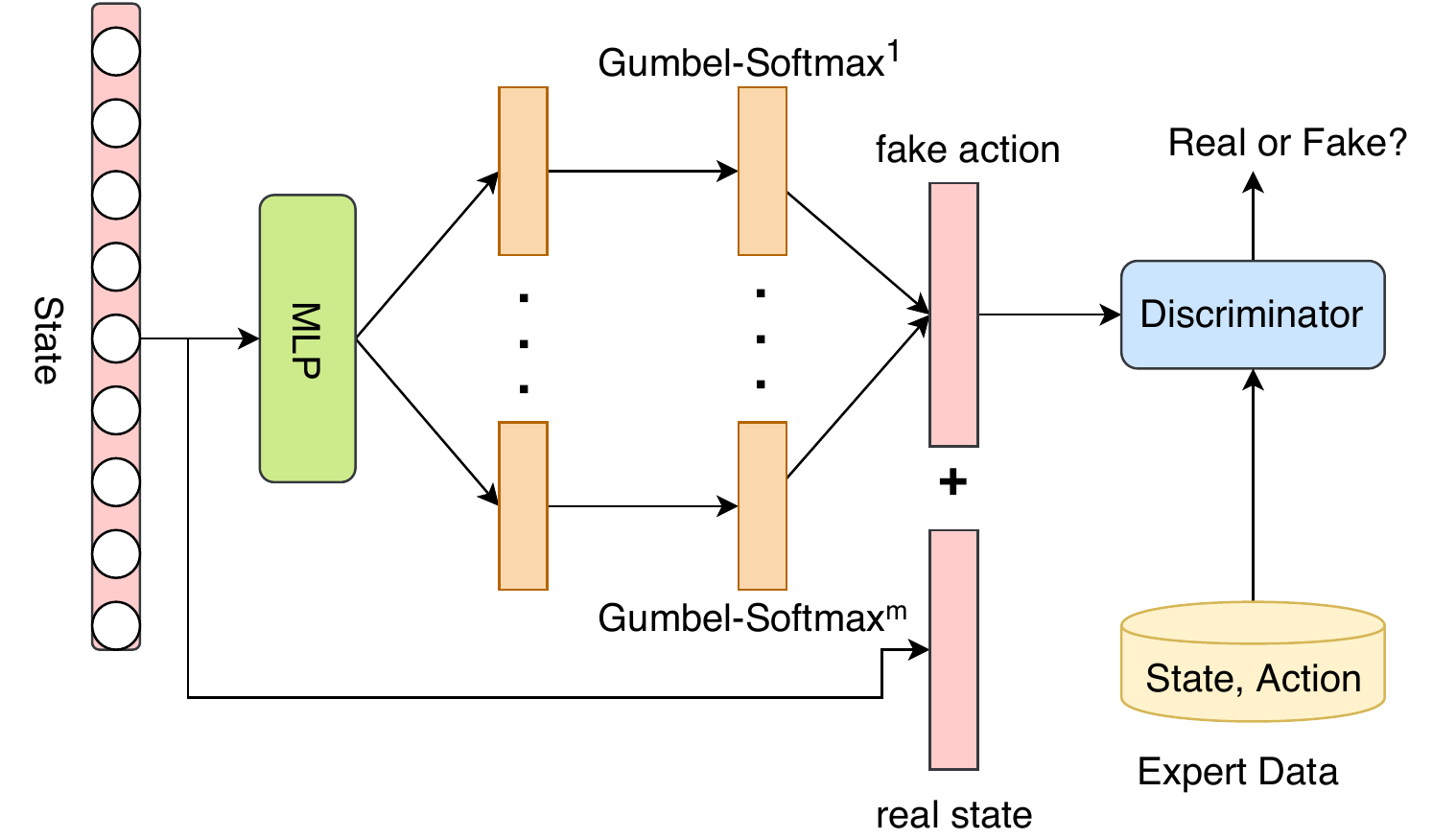}
   \caption{Architecture to approximate the dialogue policy with adversarial learning. The dialogue policy dialogue discriminator is linked to the dialogue policy through a set of Gumbel-Softmax functions\footnote{`\textbf{+} denotes the concatenation operation}.}
   \label{fig:gan_adv}
     \vspace*{-0.5\baselineskip}
\end{figure}

\noindent
\textbf{Dialogue policy} To generate dialogue actions, we employ an \ac{MLP} as the action generator $Gen_{sa}$ followed by a set of Gumbel-Softmax functions, where each function corresponds to a specific action in the atomic action space (Figure~\ref{fig:gan_adv}) and the output of each function has two dimensions. We first introduce how it works when there is only one Gumbel-Softmax function in the setting and then extend it to multiple function. The Gumbel-Max trick~\citep{gumbel1954statistical} is commonly used to draw samples $u$ from a categorical distribution with class probabilities $p$.
The process of $Gen_\theta$ can be formulated as follows:
\begin{align}
p & =  \text{MLP}(s) \\
u & = one\_hot (\argmax_i [g_i + \log p_i]),
\end{align}
where $g_i$ is independently sampled from Gumbel (0,1). However, the argmax operation is not differentiable, thus no gradient can be backpropagated through $u$. Instead, we can employ the soft-argmax approximation \citep{jang2016categorical} as a continuous and differentiable approximation to $argmax$ and to generate k-dimensional sample vectors below:
\begin{equation}
y_i = \frac{\exp((\log(p_i) + g_i)/\tau)}{\sum_{j=1}^k \exp((\log(p_j) + g_j)/\tau)}
\end{equation}
for $i=1,\ldots, k$. 
In practice, $\tau$ should be selected to balance the approximation bias and the magnitude of gradient variance. In our case, $p$ corresponds to the dialogue action status distribution $p(a_l^i|s)$ where $l \in \{0, \ldots, k-1\}$ and $i \in \{1, \ldots, m\}$. In our setting, $k$ is set to $2$ and each dimension denotes one specific action status, which could be $1$ if selected or $0$ if not selected. $m$ is set to the size of in the action space -- $166$. By taking into account the multiple actions, we rewrite the sampled vector $y$ as $y_l^i$ where $l$ and $i$ denote the corresponding dialogue action status and the $i_{th}$ atomic action in the action space respectively. The final combined action is:\footnote{$Dim(a_{fake})=166 * 2$.} 
\begin{equation}
a_{fake} = y_0^1 \oplus y_1^1 \oplus \ldots \oplus y_0^{166} \oplus y_1^{166}.
\end{equation}
Next, the generated action $a_{fake}$ is fed to the reward model $D_\omega$ along with the corresponding state $s$. The dialogue policy $Gen_\theta$ aims to get a higher reward signal from the discriminator $D$; the training loss function for the generator $Gen_\theta$ is:
\begin{equation}
L_{G}(\theta) = - \mathbb{E}_{s,a_{fake} \sim Gen}(D_{\omega}(s,a_{fake}))
\end{equation}
\noindent
\textbf{Dialogue reward} As to the dialogue discriminator, we build a reward model $D_\omega$ that takes as input the state-action pair $(s,a)$ and outputs the reward $D(s,a)$. Instead of using a discriminator to predict the probability of a generated state-action pair as being real or fake, inspired by Wasserstein GANs~\citep{arjovsky2017wasserstein}, we replace the discriminator model with a reward model that scores a given pair $(s,a)$.  Since the reward model's goal assigns a higher reward to the real data and a lower value to fake data, the objective can be given as the average reward it assigns to the correct classification. Given an equal mixture of real data samples and generated samples from the dialogue policy $Gen_\theta$, the loss function for the reward model $D_\omega$ is:
\begin{align}
L_{D}(\omega) =&  - \mathbb{E}_{s,a_{fake} \sim Gen_\theta}(D_\omega(s,a_{fake})) \\ 
& + \mathbb{E}_{s,a \sim data}(D_\omega(s,a))).
\end{align}
During training, the policy network and the reward model are be updated alternatively.

\subsection{PL as multi-label classification with dense layers}
We introduced \textit{DiaAdv}, which can bridge the policy network and the reward model together utilizing Gumbel-Softmax functions. A by-product of this framework is the policy network with dense layers and a set of Gumbel-Softmax functions. 
If we discard the Gumbel-Softmax functions but keep the dense layers, we obtain a new model, \emph{DiaMultiDense}, to solve the multi-label classification problem. Each dense layer corresponds to a specific dialogue action and the output of the dense layer has two dimensions denoting the two possible values for action status, \emph{selected} and \emph{not selected}. 
We expect the dense layers can extract informative information particularly for their corresponding actions and discard noisy information. 
During inference, the two possible values for the status of an action will be compared and the higher one will be the label for the current dialogue action. 
\textit{DiaMultiDense} can be regarded as a simple but efficient state de-noising method for dialogue \ac{PL} with multi-label classification.

\section{Experimental Setup}
\label{sec:experiments}
\noindent
\textbf{MultiWOZ Datasset}~\citep{budzianowski2018multiwoz} is a multi-domain dialogue dataset with $7$ distinct domains\footnote{Attraction, Hospital, Police, Hotel, Restaurant, Taxi, Train}, and $10,438$ dialogues. The main used scenario is a dialogue agent is trying to satisfy the tourists' demands such as booking a restaurant or recommending a hotel with specific requirements. Each dialogue trajectory is decomposed into a set of state-action pairs with the same \ac{TDS} that is used for training. In total, we have $56,700$ dialogue state-action pairs in the training set, with $7,300$ in the validation set, and $7,300$ in the test set.
\\

\noindent
\textbf{Baselines}
Three types of baselines are explored: \\ 
\textbf{(B1):} \emph{Supervised Learning}, where the dialogue action selection task is regarded as a multi-label classification problem. \\
\textbf{(B2):} \emph{\acf{RL}}, where the reward function is handcrafted and defined as follows: at the end of a dialogue, if the dialogue agent accomplishes the task within $T$ turns, it will receive $T * 2$ as a reward; otherwise, it will receive $ - T$ as a penalty. $T$ is the maximum number of turns in each dialogue; we set it to $40$ in all experiments. Furthermore, the dialogue agent will receive $- 1$ as an intermediate reward during the dialogue to encourage shorter interactions. In our experiments, we used three methods, including: \textit{GP-MBCM}~\citep{gavsic2015policy}, \textit{ACER}~\citep{wang2016sample}, \textit{PPO}~\citep{schulman2017proximal}.\\
\textbf{(B3):} \emph{Adversarial learning}, where dialogue agent is trained with a user simulator, we conduct comparisons with two methods: \textit{GAIL}~\citep{gan_imitation} and \textit{GDPL}~\citep{takanobu2019guided}. The dialogue agents in \textit{GAIL} and \textit{GDPL} are both PPO agents while these two methods have different reward models. We report the performance of  \textit{ALDM}~\citep{liu2018adversarial} for completeness.

\subsection{Training setup}
\label{section:training_settings}

\textbf{DiaSeq} With respect to \textit{DiaSeq}, we use a two-layer MLP to extract features from the raw state representation. First, we sort the action order according to the action frequency in the training set. All action combinations in the dataset will be transferred to an action path based on the action order. Three special actions -- \textit{PAD}, \textit{SOA}, \textit{EOA}, corresponding to padding, start of action decoding and end of action decoding -- are added to the action space for action decoder training. We use beam search to predict the action combinations and beam size is set to $6$. The action embedding size is set to $30$; the hidden size of the GRU is $50$.

\noindent
\textbf{DiaAdv} For the policy network of \textit{DiaAdv}, a two-layer \ac{MLP} is used to extract state features followed by $166$ dense layers and Gumbel-Softmax functions consecutively. To sample a discrete action representation, we implemented the ``Straight-Through" Gumbel-Softmax Estimator~\citep{jang2016categorical}; the temperature $\tau$ for each function is set to $0.005$. As to the discriminator, a three-layer \ac{MLP} takes as input the concatenation of dialogue state and action, and outputs a real value as the reward for the state-action pair. 

\noindent
\textbf{DiaMultiDense} We reuse the policy network from \textit{DiaAdv} except the Gumbel-Softmax functions. 

\noindent
\textbf{GDPL} \citep{takanobu2019guided} is reused. The policy network and value network are three-layer \ac{MLP}s.\\
\noindent
\textbf{PPO} The policy network in PPO shares the same architecture as \textit{GDPL}. The difference is that the reward model is replaced with a handcrafted one.\\
\noindent
\textbf{GAIL} \textit{GAIL} shares the same policy network as \textit{GDPL}. The discriminator is a two-layer \ac{MLP} taking as input the state-action pair.\\
\noindent
\textbf{DiaMultiClass} The policy network is a three-layer MLP and trained with cross entropy. It has the same architecture as the policy network in \textit{GDPL}.\\
We reuse the reported performance of  GP-MBCM, ACER, and ALDM from~\citep{takanobu2019guided} since we share the same \ac{TDS} and user simulator. 
The methods based on \ac{RL} or adversarial learning are pre-trained with real human dialogues\footnote{The code of our work: \url{https://github.com/cszmli/Rethink-RL-Sup}}. 

\subsection{Evaluation metrics}
Before a conversation starts, a user goal will be randomly sampled. The sampled user goal mainly contain two parts of information. The first part is about the constraints of different domain slots or booking requirements, e.g.\ \textit{`restaurant-inform-food'=`Thai'}, \textit{`restaurant-infor-area'=`east'}, \textit{`restaurant-book-people'=4}
which means the user wants to book a table for $4$ persons to have Thai food in the east area. The information contained in the second part is about the slot values that the user is looking for, such as \textit{restaurant-request-phone=?}, \textit{`restaurant-request-address'=?}, which means the user wants to know the phone and address of the recommended restaurant.
We use \emph{Match}, \emph{Recall}, \emph{F1 score} to check if all the slot constraints and requested slot information have been satisfied. \emph{F1 score} evaluates whether all the requested information has been provided while \emph{Match} evaluates whether the booked entities match the indicated constraints. We use \emph{Average Turn} and \emph{Success rate} to evaluate the efficiency and level of task completion of dialogue agents. If an agent has provided all the requested information and made a booking according to the requirements, the agent completes the task successfully.

\section{Results and Discussion}
\label{sec:results}

\subsection{Performance of different dialogue agents}
\label{sec:automatic_results}

\begin{table}[t]
\centering
\resizebox{1.\columnwidth}{!}{
\begin{tabular}{l*{5}{c}}
\toprule
\textbf{Dialogue agent} & \textbf{Turn}  & \textbf{Match}  & \textbf{Rec} & \textbf{F1}  & \textbf{Success rate}  \\
\midrule
GP-MBCM          & \phantom{1}2.99    & 0.44   & -- & 0.19  & 28.9\\
ACER             & 10.49    &0.62    & -- & 0.78  & 50.8\\
PPO (human)       & 15.56  &0.60   & 0.72 & 0.77  & 57.4\\
ALDM             & 12.47    &0.69    & --   & 0.81  & 61.2\\
GDPL      & \phantom{1}7.80  &0.81    & 0.89 & 0.87  & 81.7\\
GAIL    & \phantom{1}7.96 & 0.81  & 0.87 & 0.86 & 80.5 \\
\midrule
DiaMultiClass   & 12.66 & 0.58  & 0.71 & 0.79 & 57.2 \\
\midrule
DiaMultiDense   & \phantom{1}9.33 &0.85 &0.94 &0.87 & 86.3\rlap{\smash{$^*$}}  \\
DiaSeq          & \phantom{1}9.03 & 0.81  & 0.88 & 0.85  & 81.6\\
DiaAdv    &\phantom{1}8.80  &0.85   &0.94   &0.85  &87.4\rlap{\smash{$^*$}} \\
\bottomrule
\end{tabular} 
}
\caption{The performance of different dialogue agents, which is calculated based on the average results by running each method $5$ times. * indicates statistically significant improvements ($p<0.005$) using a paired t-test over the \textit{GDPL} success rate and the proposed methods.} 
  \vspace*{-0.5\baselineskip}
\label{Table:results_full}
\end{table}

Tab.~\ref{Table:results_full} shows the performance of different dialogue agents.
With respect to success rate, \textit{DiaAdv} manages to achieve the highest performance by $6\%$ compared to the second highest method \textit{GDPL}. However, \textit{DiaAdv} is not able to beat \textit{GDPL} in terms of average turns. A possible reason is that \textit{GDPL} can generate more informative and denser dialogue action combinations. With a user simulator in the training loop, the dialogue agent can explore more unseen dialogue states in the dataset. Furthermore, the same user simulator will be used to test the dialogue agent and the dialogue agent will definitely benefit from what he has explored in the training stage. However, more informative and denser responses will not guarantee all the users' requirements will be satisfied and this will lead to a lower \emph{Match} score as shown in Tab.~\ref{Table:results_full}.  

As to \textit{DiaSeq}, it can achieve almost the same performance as \textit{GDPL} from different perspectives while \textit{GDPL} has a slightly higher \emph{F1} score. However, the potential cost benefits of \textit{DiaSeq} are huge since it does not require a user simulator in the training loop. The training of \textit{DiaSeq} is well-understood and we can get rid of tuning the sensitive parameters in \ac{RL} and Adversarial Learning. To sum up, \textit{DiaSeq} is far more cost-efficient solution.

Another supervised learning method, \emph{DiaMultiDense} achieves remarkable performance with respect to different metrics. Compared to the traditional solution \textit{DiaMultiClass}, joining of dense layers as in \emph{DiaMultiDense} brings a huge performance gain; it manages to beat \textit{DiaMultiClass} on all the metrics. And it achieves higher \emph{F1} score than \textit{DiaAdv}. Since the only difference between \emph{DiaMultiDense} and \textit{DiaMultiClass} is that we replace the last layer of \textit{DiaMultiClass} with a stack of dense layers, the change in the number of parameters may lead to the performance gap. We report the number of parameters of three supervised learning methods in Tab.~\ref{Table:results_params}. \emph{DiaMultiDense} achieves the highest performance among these three methods while using the fewest parameters. We believe the dense layers have been trained to filter noisy information from the previous module and the final classification can benefit from the high-quality information flow.
\begin{table}[t]
  \centering
  \resizebox{1.\columnwidth}{!}{
\begin{tabular}{l*{3}{c}}
\toprule
\textbf{Dialogue agent} &  DiaSeq   &  DiaMultiClass  &  DiaMultiDense \\
\midrule             
\#\textbf{Parameters}   &251,000 &184,000 &133,000\\
\bottomrule
\end{tabular} 
}
\caption{Total number of parameters for supervised learning models.} 
\label{Table:results_params}
\end{table}

\begin{table*}
  \centering
  \resizebox{\linewidth}{!}{
  \begin{tabular}{ cc*{12}{c}}
    \toprule
  \multirow{2}{*}{\diagbox{Dataset}{\makecell{Agent}} }  & \multicolumn{2}{c}{ DiaMultiClass} & \multicolumn{2}{c}{DiaSeq} & \multicolumn{2}{c}{DiaMultiDense}&
  \multicolumn{2}{c}{GDPL} &
  \multicolumn{2}{c}{DiaAdv} &
  \\
  \cmidrule(r){2-3}
  \cmidrule(r){4-5}
  \cmidrule(r){6-7}
  \cmidrule(r){8-9}
  \cmidrule(r){10-11}
&\makecell{Turn} & \makecell{Success rate} &\makecell{Turn} &  \makecell{Success rate}
&\makecell{Turn} & \makecell{Success rate}
&\makecell{Turn} & \makecell{Success rate}
&\makecell{Turn} & \makecell{Success rate}\\
\midrule
\makecell{MultiWOZ (0.1)} & 17.14  & 31.7 & 10.77 & 70.4 &18.36  & 27.0 &9.21 &21.2 &16.80 &37.2\\
\makecell{MultiWOZ (0.4)} & 12.56 & 59.0  & 9.99 & 75.5   & 10.76   & 79.4  &8.49 &68.0 &9.90 &81.6\\
\makecell{MultiWOZ (0.7)} & 13.1  & 53.6      & 9.35  & 77.2   & 10.02   &85.1 &8.10 &73.3  &9.30 &87.0\\
    \bottomrule
  \end{tabular}
  }
\caption{The performance of different dialogue agents with different amounts of expert dialogues. We only report \emph{Average Turn} and \emph{Success rate} here due to limited space.} 
\label{Table:results_diffsize}
\end{table*}

\subsection{User experience evaluation}
\begin{table}[t]
  \centering
  \resizebox{.8\columnwidth}{!}{
\begin{tabular}{l*{3}{c}}
\toprule
\textbf{Dialogue pair} & \textbf{Win}  & \textbf{Loose}  & \textbf{Tie} \\
\midrule             
DiaMultiDense vs. GDPL   &42 &50 & \phantom{1}8\\
DiaSeq vs. GDPL   &50 &44 &\phantom{1}6\\
DiaAdv vs. GDPL   &39 &51 &10\\
\bottomrule
\end{tabular} 
}
\caption{Human evaluation results.} 
\vspace*{-0.5\baselineskip}
\label{Table:results_human}
\end{table}
Automatic metrics can only capture part of the performance difference between different dialogue agents. For example, we use success rate to reflect the level of task completion and use turn number to represent the efficiency of dialogue agents. However, the final goal of a \ac{TDS} is to assist real users to complete tasks. To fully evaluate system performance while interacting with real users, we launch an evaluation task on Amazon Mturk to rate the user experience with the proposed dialogue systems. For each evaluation task, we will first present an Mturk worker with a randomly sampled user goal, which contains the constraints about specific domain slots and some slot information that the user is looking for. In the next step, according to the sampled goal, two generated dialogues from two different dialogue agents are shown to the worker. The worker needs to pick up the dialogue agent that provides a better user experience. Different factors will be taken into account, such as response quality, response naturalness, how similar it is compared to a real human assistant. If the worker thinks two dialogue agents perform equally good/bad or it's hard to distinguish which one is better, the option `Neutral' can be selected. Four dialogue agents are evaluated: \textit{GDPL},  \textit{DiaSeq}, \emph{DiaMultiDense} and \textit{DiaAdv}, and there are three comparison pairs \textit{DiaMultiDense-GDPL}, \textit{DiaSeq-GDPL}, \textit{DiaAdv-GDPL} since \textit{GDPL} is regarded as the \ac{SOTA} method. Each comparison pair has 100 dialogue goals sampled and 200 corresponding dialogues from two different dialogue agents. All the dialogue actions in the dialogue turns are translated into human readable utterances with the language generation module from ConvLab \citep{lee2019convlab}. Each dialogue pair is annotated by three Mturk workers. The final results are shown in Tab.~\ref{Table:results_human}.

The method \textit{DiaAdv} can be regarded as an extension of \textit{DiaMultiDense} by adding a classifier to provide a stronger training signal. According to the results from Section~\ref{sec:automatic_results}, these two methods do improve the success rate of dialogue agents. However, as shown in Tab.~\ref{Table:results_human}, while the success rate improves, the user experience degrades. According to Tab.~\ref{Table:results_full}, \textit{GDPL} and \textit{DiaAdv} have similar \emph{F1} scores but the \textit{DiaAdv} has a higher \emph{Recall} value; this means that \textit{DiaAdv} achieves a lower \emph{Precision}. The unnecessary information mixed in the system response annoys users and results in a lower user experience. Given the relatively large difference in terms of success rate, the trade-off between success rate and user experience should be carefully examined. From another perspective, it is understandable that \textit{GDPL} can provide a better user experience because a pre-designed user simulator is involved and the discriminator will encounter more diverse state-action combinations that are not seen in the training data.  In contrast, the discriminator in \textit{DiaAdv} only has access to the training data and this limits its judging ability. This does not imply that having a user simulator in the loop is essential to provide high quality user experience: \textit{DiaSeq}, which is a completely supervised learning method, outperforms \textit{GDPL}. 

\subsection{Discussion}
 
\textbf{How many expert dialogues are enough to train a dialogue agent with supervised learning?}
One motivation for dropping supervised learning and employing \ac{RL} methods in \ac{TDS} is that building high-quality conversational datasets is expensive and time-consuming. In contrast, training  dialogue agents with a user-simulator is cheaper and affordable in many cases. Since we have no control on how much domain knowledge should be involved to build a user-simulator, we are not able to measure the expense of a reliable user-simulator. However, we can conduct an experiment to show how many real human dialogues are required to train a high-quality dialogue agent. 

Based on the original MultiWoZ dataset, we build three smaller subsets: \emph{MultiWoZ(0.1)}, \emph{MultiWoZ(0.4)}, \emph{MultiWoZ(0.7)} by only keeping $10\%$, $40\%$, and $70\%$ dialogue pairs from the original dataset, respectively. We retrain \textit{DiaMultiClass}, \textit{GDPL}, \textit{DiaAdv}, \textit{DiaMultiDense}, \textit{DiaSeq} and report the performance in Tab.~\ref{Table:results_diffsize}.
With respect to supervised learning agents, with only $10\%$ expert dialogue pairs, \textit{DiaMultiClass} gets half the success rate compared to the original performance (Tab.~\ref{Table:results_full}). By adding $30\%$ more dialogue pairs to the training set, \textit{DiaMultiClass} can achieve the same performance $59\%$ with the original success rate $57.2\%$. Beyond this, \textit{DiaMultiClass} does not benefit from the increase in expert dialogues and starts to fluctuate between $55\%$ and $59\%$. In contrast, \textit{DiaSeq} can achieve higher performance when there are only $10\%$ expert dialogue pairs and the success rate increases with the number of available expert dialogues. \textit{DiaMultiDense} achieves the best performance with the same amount of expert dialogues compared to the other two supervised learning methods. The performance difference among the three supervised learning methods shows that the method itself is the main factor to influence the performance rather than the number of available expert dialogues in the given dialogue environment. To some extent, traditional \textit{DiaMultiClass} does not exert the potential of a given dataset to the fullest in dialogue \ac{PL}.

\smallskip\noindent%
\textbf{Can adversarial learning eliminate expert dialogues?}
As can be concluded from Tab.~\ref{Table:results_diffsize}, \textit{GDPL} and \textit{DiaAdv} managed to improve the performance with the increasing number of expert dialogues. 
\textit{GDPL} and \textit{DiaAdv} have the reward models that are supposed to distinguish real dialogue pairs from the machine-generated ones. By observing more expert dialogues, the reward model can provide a dialogue policy with more reliable and consistent updating signals.
\begin{figure}[ht]
\centering
   \includegraphics[clip, width=1.0\columnwidth]{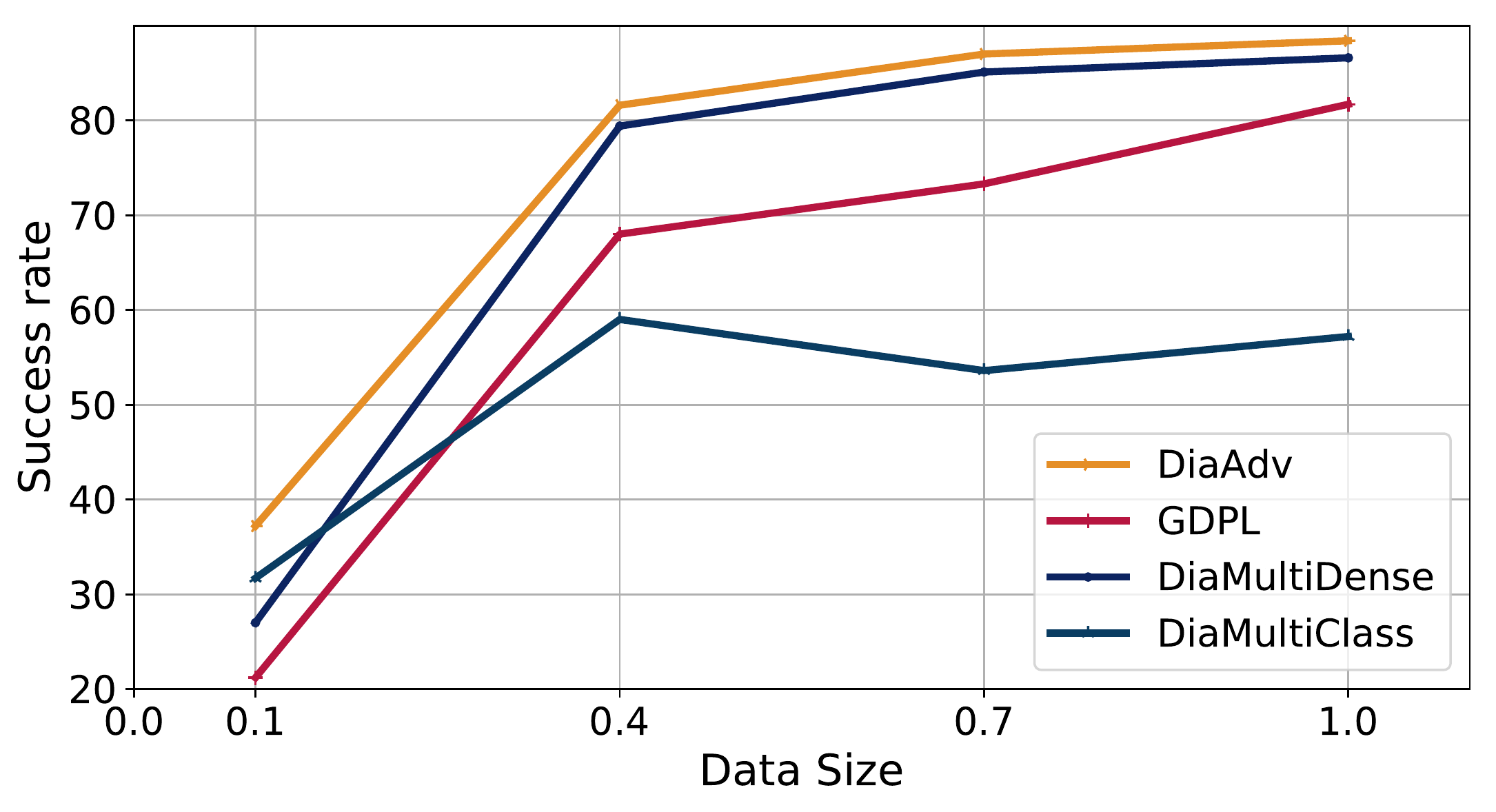}
    \vspace*{-\baselineskip}
   \caption{The performance gain between the pre-trained and their corresponding adversarial learning models with different amounts of expert dialogues.}
   \label{fig:datasize}
\end{figure} 
Figure~\ref{fig:datasize} shows the success rate gain by applying adversarial learning methods to the corresponding pre-trained models \footnote{\textit{DiaAdv} is the adversarial extension of \textit{DiaMultiDense} while \textit{GDPL} is the adversarial extension of \textit{DiaMultiClass}.}. 
When the success rates of \textit{DiaMultiClass} with MultiWoZ(0.4) and MultiWoZ(1.0) are both around $60\%$, deploying \textit{GDPL} manages to bring $10\%$ performance gain. The performance difference can be caused by the improved quality of the reward model. Conversely, if the reward model has no access to sufficient amount of expert behaviors, it has little clue how the expert dialogues should look like. This can lead to poor reward signals for the policy network. We can see it in the case of \textit{GDPL} that the success rate drops to $21\%$ while the pre-trained model can achieve $31\%$ success rate on MultiWoZ(0.1). 
The performance gain between \emph{DiaMultiDense} and \textit{DiaAdv} is not so remarkable with respect to success rate compared to the gain between \textit{DiaMultiClass} and \textit{DiaAdv}. However, \textit{DiaAdv} does help to reduce the dialogue turns while improving the success rate as shown in Tab.~\ref{Table:results_diffsize}. We can regard \textit{DiaAdv} as a promising method to fine-tune the \emph{DiaMultiDense} to explore more potential dialogue states.

\smallskip\noindent%
\textbf{How sensitive are adversarial learning to pre-trained dialogue policy?}
We explore how pre-trained dialogue policies affect the final performance of adversarial learning based dialogue agents. 
We first use supervised learning to pre-train the dialogue policies of \textit{GDPL} and \textit{DiaAdv} respectively with different training epochs. 
\begin{figure}[ht]
\centering
   \includegraphics[clip, width=1.0\columnwidth]{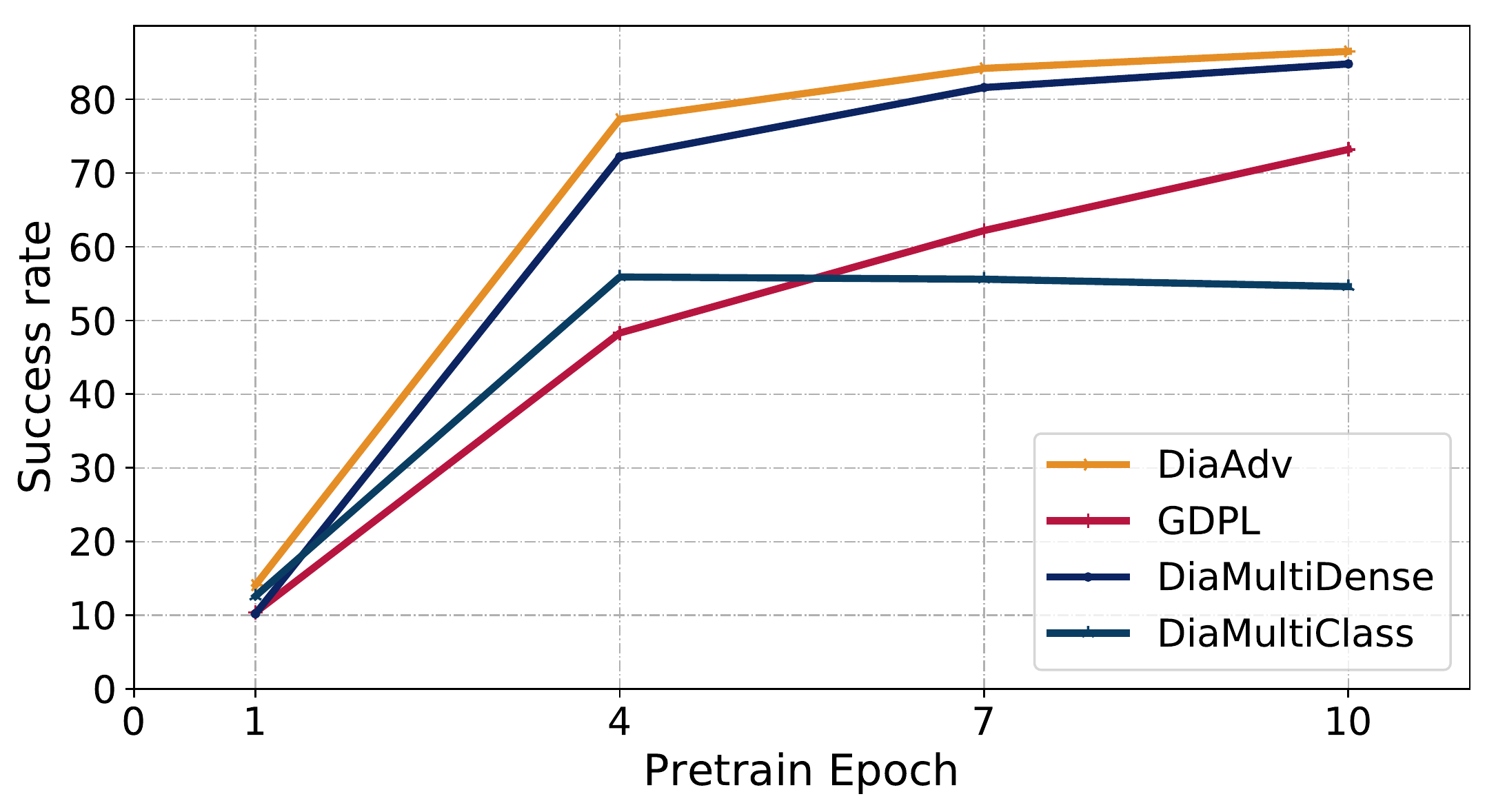}
    \vspace*{-\baselineskip}
   \caption{The performance gain between the pre-trained and their corresponding adversarial learning models with different amounts of pre-taining epochs.}
   \label{fig:diffpre}
\end{figure} 
As shown in Figure \ref{fig:diffpre}, the performance gain between the pre-trained dialogue policy and the corresponding adversarial are limited. 
With respect to \textit{GDPL}, it even degenerates the original performance of the pre-trained policy when the starting points are relatively low. 
In other words, the main contributions to the adversarial dialogue agents come from the supervised learning stage; it is challenging for the dialogue agents to achieve the same performance without a promising pre-trained dialogue policy.  

\section{Conclusion}
In this work, we proposed two supervised learning approaches and one adversarial learning method to train the dialogue policy for \acp{TDS} without building user simulators. The proposed methods can achieve state-of-the-art performance suggested by existing approaches based on \acf{RL} and adversarial learning. However, we have demonstrated that our methods require fewer training efforts, namely the domain knowledge needed to design a user simulator and the intractable parameter tuning for \ac{RL} or adversarial learning. Our findings have questioned if the full potential of supervised learning for dialogue \acf{PL} has been exerted and if \ac{RL} methods have been used in the appropriate \ac{TDS} scenarios. 

\clearpage

\bibliographystyle{acl_natbib} 
\bibliography{bibliography}

\end{document}